# Towards Effective Image Forensics via A Novel Computationally Efficient Framework and A New Image Splice Dataset

Ankit Yadav[1,*], Dinesh Kumar Vishwakarma[2]

Biometric Research Laboratory, Department of Information Technology, Delhi Technological University, Bawana Road, Delhi-110042, India

ankit4607@gmail.com[1,**], dvishwakarma@gmail.com[2]

**ABSTRACT**

Splice detection models are the need of the hour since splice manipulations can be used to mislead, spread rumors and create disharmony in society. However, there is a severe lack of image-splicing datasets, which restricts the capabilities of deep learning models to extract discriminative features without overfitting. This manuscript presents two-fold contributions toward splice detection. Firstly, a novel splice detection dataset is proposed having two variants. The two variants include spliced samples generated from code and through manual editing. Spliced images in both variants have corresponding binary masks to aid localization approaches. Secondly, a novel *Spatio-Compression Lightweight Splice Detection Framework* is proposed for accurate splice detection with minimum computational cost. The proposed dual-branch framework extracts discriminative spatial features from a lightweight spatial branch. It uses original resolution compression data to extract double compression artifacts from the second branch, thereby making it 'information preserving.' Several CNNs are tested in combination with the proposed framework on a composite dataset of images from the proposed dataset and the CASIA v2.0 dataset. The best model accuracy of 0.9382 is achieved and compared with similar state-of-the-art methods, demonstrating the superiority of the proposed framework.

**Statements and Declarations**

**Data Availability:** The datasets generated during or analyzed during the current study are available through an online web repository via the following weblinks: https://www.kaggle.com/datasets/divg07/casia-20-image-tampering-detection-dataset
**Funding:** No funding was received to assist with preparing this manuscript.
**Conflict of Interest:** The authors declare that they have no known competing financial interests or personal relationships that could have appeared to influence the work reported in this paper.
**Ethical Approval:** Not Applicable.
**Patient Consent Statement**: Not Applicable.
**Permission to Reproduce Material from Other Sources**: Not Applicable.
**Author Contributions:** Ankit Yadav - Software, Validation, Investigation, Data Curation, Writing Original Draft, Visualization. Dinesh Kumar Vishwakarma - Conceptualization, Methodology, Formal Analysis, Resources, Writing – Review & Editing, Supervision, Project Administration, Funding Acquisition.

## 1 Introduction

The constantly growing popularity of social media platforms in the last decade has created vast volumes of data per user. As many as 3.96 billion social media users are active, double the figure at the beginning of 2015 [1]. Such widespread use of social media has resulted in a constantly increasing everyday data explosion, hence the term Big Data. Countless videos, images, and text data are being created and uploaded through social media users, with an average of 8.8 social media accounts per person [1].

* Corresponding Author
** Corresponding Author Email

The growing popularity of online presence is also paralleled by the growth of several multimedia manipulation tools and approaches [2, 3]. Computer tools such as Adobe Photoshop and an endless supply of image/video editing mobile applications provide the end-user with readymade and easy-to-use manipulation capabilities. While most of these editing applications are well intended for creating visual enhancements and providing entertaining and humorous content, these tools are easily repurposed for ill-intentional deeds.

Images and video manipulations pose a more significant threat to society than textual misinformation since visual data is more believable. An image or a video of any wrongdoing is instantly accepted as the truth. And yet, today's manipulation tools make it child's play to maliciously manipulate an image/video and cause harm to the reputation of an individual or an organization. Some harmful effects of visual manipulations are defamation, fraud (monetary or identity-wise), spreading fake news and rumors, misdirection, swaying public opinion for political gain, etc.

'Image Splicing' is one such image/video manipulation where an object from one image is pasted onto another image. Such modification to multimedia data can easily be used to mislead from the truth and cause harm to a given person or organization. Facial splicing can cause identity misdirection, and an innocent person can easily be made to look guilty of wrongdoings. Hence, it is crucial to build robust splice detection approaches capable of detecting manipulations in visual data [4]. An additional consideration for splice detection is that since most visual data is compressed to save storage space, splice detection through compression artifacts can form a significant starting point toward robust splice detection. Specifically, since most compressed images are created using the famous JPEG compression, its artifacts in spliced images form the basis of the novel contribution proposed in this paper.

The main contributions of this work are: -

- Proposed a novel splice detection dataset – *BiometricLab-DTU-Splice Dataset* is proposed. The proposed dataset has two variants. The first variant is autogenerated from code, while the second contains handmade spliced samples. Binary masks are available in both variants. Given the size of existing small-scale splice datasets, the proposed dataset is a significant addition to the splice detection research arena.
- Proposed a novel lightweight, dual-branch, information-preserving, spatial-compression modal splice detection framework is proposed to detect spliced jpeg images while restricting the computational complexity to a small fraction of the usual computational cost in the context of deep learning.
- The proposed model contains a novel 'spatial branch' to extract discriminative spatial information for detecting image splicing. Transfer learning is used to leverage the strong classification capabilities of deep models in the spatial domain at minimal computational costs.
- The proposed model contains a novel information preserving 'compression branch' which uses original resolution compression data to extract double compression artifacts from spliced jpg.
- The proposed splice detection framework is designed to be extremely lightweight.

- Experimentations with several variants of the proposed spliced detection framework and comparisons with existing splice detection methods prove the potency of the proposed splice detection framework at minimum computational costs.

The organization of this research manuscript is as follows: Section 2 discusses several existing state-of-the-art splice detection/localization algorithms and highlights the research gaps that are the motivation behind the design of the proposed splice detection framework. Section 3 explains the design principles of the proposed splice detection framework and details the specifications of the spatial branch, compression branch and the final model. Section 4 dives into the experimentation details of this research manuscript. Firstly, it introduces the novel BiometricLab-DTU Splice dataset variants (automatic and manual) and specifies a modified CASIA v2.0 dataset. Secondly, several variants of the proposed splice detection framework are implemented and evaluated on the datasets. Thirdly, the proposed splice detection framework models are compared against existing state-of-the-art methods. An ablation study is also conducted to verify the necessity of a dual-branch framework. Section 5 presents the conclusion highlighting the novel contributions of this research manuscript.

## 2 Related Work

Several significant contributions have been made toward splice detection and localization in images and videos. For example, Chen et al. [5] proposed a novel splice detection method by highlighting distinct shapes of intensity-gradient bivariate histograms near edge regions of a spliced image caused by non-linear camera response functions (CRF). Bondi et al. [6] input image patches to detect source camera attributes of each pixel within an image patch. A CNN extracts source camera features, while a clustering algorithm estimates the tampering mask. Pomari et al. [7] highlight regions of illumination inconsistencies to detect splicing. Specifically, a pre-trained ResNet50 model without the classifier layer is used and fine-tuned to extract 'deep splicing features' (DSF) from the illuminant maps of an input image. Verde et al. [8] propose a novel contribution for temporal splice detection in videos. The authors target identifying the fusion of video segments with distinct characteristics, such as different recording devices or broadcasting channels, by training two CNNs to learn video codec and coding quality features. Salloum et al. [9] use fully convolutional neural networks to localize image splicing. The authors use two networks for multi-task learning that learn ground truth masks and the boundary of spliced regions, respectively. Cun et al. [10] propose the novel 'Semi-Global Network' that utilizes global features from the entire image and local patch-level features for splice localization. Local features correspond to spliced edges, while global information contains semantic and illumination information. Fully Connected Conditional Random Fields are used as a post-processing step. Liu et al. [11] propose a novel 'Fusion-net' architecture that combines several deep CNNs trained to detect image splicing based on different attribute artifacts, such as noise, compression, etc., for splice localization.

Mazumdar et al. [12] utilize pairwise facial illumination maps to train a pair of CNNs in a Siamese network setup. Such an architecture learns discriminative features of facial illuminant pairs from the same or distinct illumination environments. Later, one of the trained CNN can be used as a feature extractor for facial splicing detection. Bi et al. [13] introduce a novel 'RRU-Net' that makes better use of contextual spatial information within an image without the requirement of any pre/post-processing. The novel architecture prevents the gradient degradation problem common in massive networks and proves effective for image splice detection. Deng et al. [14] utilize a novel 'MSD-Net' architecture that learns multi-scale features from DCT coefficient histogram features from jpg images. Specifically, the proposed network extracts feature at three scales using three interconnected networks to classify the input image block as original or spliced. Horváth et al. [15] detect splice forgery in satellite images using deep belief networks containing two stacked restricted Boltzmann machines. The deep belief network inputs patches from original satellite images and produces reconstructed patches and normalized error maps computed from input and reconstructed patches. Xiao et al. [16] introduce a novel 'CR2Net' that extracts distinct image properties from original and forged regions using a coarse CNN (C-CNN) and a refined CNN (R-CNN) at different scales. Forged regions are generated using diluted adaptive clustering. Wang et al. [17] fuse weighted YCbCr, edge, and camera PRNU features through novel weight combination modules for image splice detection. The fusion parameters are auto-tuned during backpropagation to ensure the best combination of features is utilized for spliced forgery detection. Liu et al. [18] perform feature learning through a deep fusion network that distinguishes differences in noise and compression information of original and spliced samples.

Based on the existing literature on splice detection, two crucial inferences are drawn: -

- The number and size of splice detection datasets are minimal in deep learning. Training deep architectures on such small-scale datasets inevitably leads to overfitting.
- Several contributions prepare their splice data to evaluate their proposed splice detection algorithms.

Table 1 clearly illustrates the lack of large-scale publicly available splice detection/localization datasets. Most splice detection methods either synthesize new datasets or use the existing small-scale splice datasets (only CASIA v2.0 contains 10000+ total samples). The characteristics of these publicly available splice datasets are discussed in greater detail in Table 4.

Table 1 Most Splice Detection/Localization contributions either prepare a new dataset for proposed model evaluation or use small-scale publicly available splice datasets

| Ref. | Self-Prepared Dataset | Publicly Available Dataset | Total Samples in Publicly Available Splice Datasets |
|---|---|---|---|
| [5] | ✓ | ✓ | 363 (Columbia) |
| [6] | ✓ | | - |
| [7] | | ✓ | 613 (Columbia + DSO-I + DSI-I) |
| [8] | ✓ | | |
| [9] | | ✓ | 14948 (CASIA v1 + CASIA v2 + Columbia + DSO-I + DSI-I) |
| [10] | | ✓ | 613 (Columbia + DSO-I + DSI-I) |
| [11] | ✓ | | - |
| [12] | | ✓ | 250 (DSO-I + DSI-I) |
| [13] | | ✓ | 14698 (CASIA v1 + CASIA v2 + Columbia) |
| [14] | ✓ | ✓ | 100 (Florence) |
| [15] | ✓ | | - |
| [16] | | ✓ | 3196 (CASIA v2 + Columbia + Forensics) |
| [17] | | ✓ | 14335 (CASIA v1 + CASIA v2) |
| [18] | | ✓ | 803 (Columbia + Realistic Image Tampering) |

Such data-oriented limitations in the splice detection research arena provide strong motivation to set the following experimental goal:

- Propose a comprehensive splice detection dataset
- Propose a lightweight splice detection framework to handle small-scale splice datasets without overfitting the training data. The design principles of the proposed framework are discussed in the next section.

## 3 Proposed Splice Detection Framework

The proposed splice detection framework is discussed in this section. Firstly, the design principles that govern the proposed splice detection framework are elaborated. Secondly, the spatial branch characteristics are highlighted. Thirdly, the highlights of the compression branch are specified. Lastly, the fusion of features from both branches is discussed to describe the final model.

### 3.1 Design Principles

The design of the proposed splice detection framework is motivated by several factors. Firstly, while spatial features are the most common type of deep features extracted, they do not provide the most discriminative information about spliced samples. The literature review in Section 2 suggests splice detection in different modalities, including frequency domain, noise domain, etc., has proven effective. Secondly, the small number and size of publicly available splice datasets (Table 4) provide a clear incentive to avoid heavy deep learning architectures having millions of trainable parameters. A deep model that is too complex for a given dataset will overfit and memorize the training samples. Thirdly, since deep learning architectures require fixed-size inputs, images are mainly resized to smaller resolutions, and hence there is 'information loss' due to the reduction of high-dimensional images.

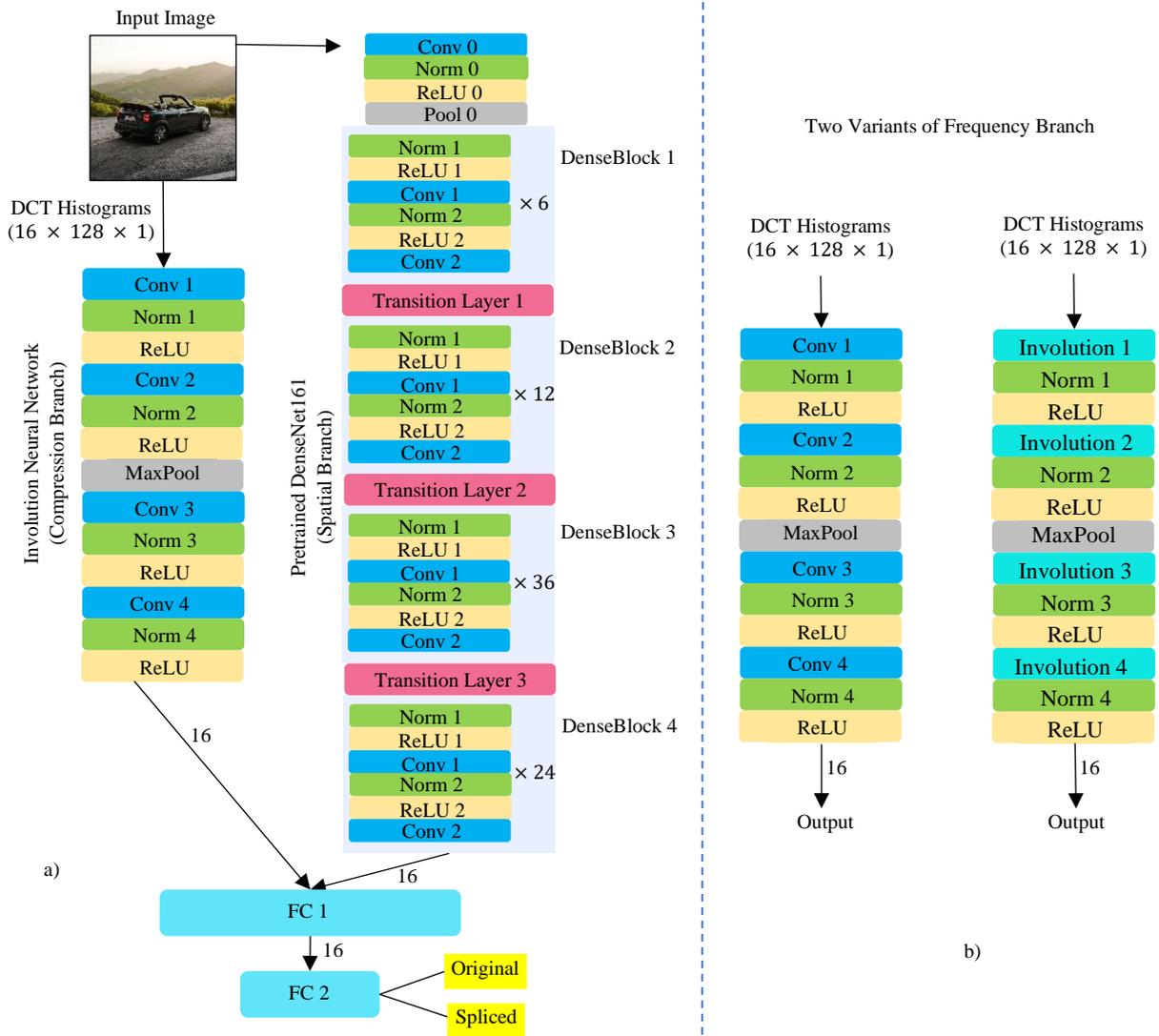

Fig. 1 a) DenseNet-CNN variant of the proposed Splice Detection Framework containing a pre-trained DenseNet161 for the spatial branch and a Convolution Neural Network (CNN) for the compression branch. b) The frequency branch of the proposed splice detection framework has two variants based on convolution and involution operators.

Hence, a splice detection framework is proposed with the following novel characteristics:

- **Dual-branch for Multi-Modal Feature Learning** – Different modalities besides the spatial domain have proven effective for splice detection. This proposed framework combines the spatial domain with a 'compression branch' that learns discriminative compression artifacts indicating image splicing.
- **Information Preserving** – The compression branch of the proposed framework extracts compression artifacts from original resolution image data. Hence, no information is lost due to resizing.
- **Lightweight** – The design of the proposed splice detection framework restricts the number of trainable parameters (Table 7) in the context of deep learning. None of the proposed framework variants have more than 100,000 trainable parameters, whereas standard deep networks can easily have up to millions of trainable parameters. Fewer trainable parameter leads to significantly lower computational cost for the proposed splice detection framework.
- **Futuristic** – A novel framework is proposed instead of building a fixed architecture for splice detection. The proposed framework supports a variety of existing deep architectures and is also capable of utilizing novel ideas from future research. This plug-and-play characteristic ensures that the proposed framework stays relevant for years.

Table 2 Several Deep Architectures are used for the spatial and compression branch of the proposed framework.

| Ref | Model | Short Form | Branch | Input Size | Key Idea |
|---|---|---|---|---|---|
| [19] | VGG16 | VGG | Spatial | 256 x 256 x 3 | Small convolutional kernels |
| [20] | GoogleNet | - | Spatial | 256 x 256 x 3 | Inception module, parallel convolution |
| [21] | ResNet18 | ResNet | Spatial | 256 x 256 x 3 | Skip Connections |
| [22] | DenseNet161 | DenseNet | Spatial | 256 x 256 x 3 | Feature Reuse |
| [23] | Vision Transformer | ViT | Spatial | 384 x 384 x 3 | Self-Attention on Images |
| - | Convolutional Neural Network | CNN | Compression | 16*128*1 | Channel-Specific and Spatial-Agnostic |
| [24] | Involution Neural Network | INN | Compression | 16*128*1 | Spatial-Specific and Channel-Agnostic |

### 3.2 Spatial Branch

As the name suggests, the spatial branch extracts spatial features from input image samples. Several convolutional neural network architectures have proven highly effective for image classification problems. The spatial branch is designed to use transfer learning via deep networks pre-trained on the ImageNet dataset to inherit the proposed framework's lightweight and' futuristic characteristics. Designing the spatial branch in this manner has the following advantages. Firstly, several architectures ( [19], [20], [21], [22]) that have proven to possess high image classification capabilities can be leveraged to extract discriminative spatial features for splice detection. Secondly, transfer learning ensures that the spatial branch remains 'lightweight,' i.e., only the last layer is trainable. Deep architectures of any size can be used for the spatial branch as long as it is pre-trained on ImageNet, and all layers are frozen except the last layer.

The spatial branch's last layer in the proposed framework is trainable and modified to produce features of 16 dimensions. Fig. 1 shows a variant of the proposed splice detection framework that uses an ImageNet pre-trained 'Vision Transformer' [23] in the spatial branch. Table 2 shows several pre-trained deep architectures used for the spatial branch of the proposed spliced detection framework. All models receive color images as input. It is restated that such a design enjoys strong classification capabilities in the spatial domain while staying lightweight regardless of the size of the deep architecture used.

### 3.3 Compression Branch

Several existing works highlight the presence of distinct compression artifacts in spliced jpg images [25], [26]. Specifically, an original jpg image (without splicing) is compressed once. However, a spliced image containing an object pasted from another image undergoes a second jpg compression. This dual compression leaves distinct artifacts in DCT coefficient histograms. The DCT coefficients include 1 DC and 63 AC coefficients. Different works have considered different AC coefficients in zig-zag order with different histogram ranges. [25] highlights the artifacts by using 9 DCT coefficients (zig-zag order) and a histogram range of $[-5,5]$ to constitute features of size $99 \times 1$. Similarly, [26] considers 9 DCT coefficients with a histogram range of $[-50, 50]$ to formulate a feature size of 909.

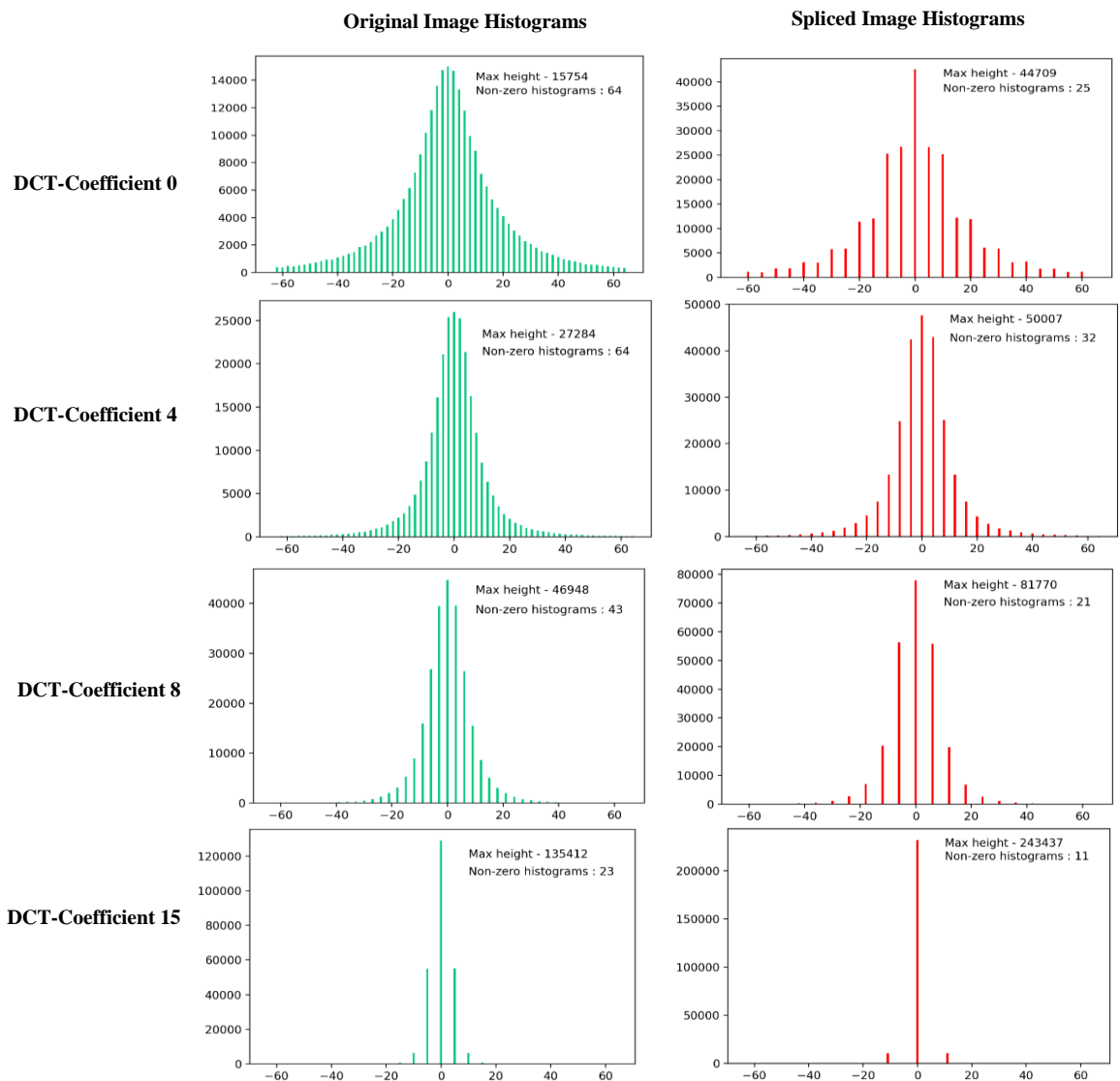

Fig. 2 This figure shows the double compression artifacts in DCT coefficient histograms of spliced jpg images. Four coefficient plots (0, 4, 8 and 15th AC coefficient in zig-zag order) are drawn for an original (green) and spliced (red) image. The histograms from original and spliced images obtain distinct shapes with different maximum values and different non-zero histogram bins up to 16 DCT AC coefficients in zig-zag order

Visual analysis is conducted on several original-spliced image pairs, and histograms are plotted to evaluate the ideal range of features. Fig. 2 shows the histograms from an original (green) and corresponding spliced (red) image. Plotting the zeroth, fourth, eighth, and fifteenth AC coefficient (zig-zag order) demonstrates the presence of distinct maximum values and a varying number of non-zero histogram bins in the range of $[-63, 64]$ between the original and spliced image. Hence, 16 AC zig-zag coefficients (0 to 15) are selected, and the histogram range is set to $[-63, 64]$. The input size for the compression branch is set to $16 \times 128 \times 1$.

The histogram data input is extracted from original resolution images; hence, the compression branch is 'information preserving,' i.e., there is no loss of information from input resizing. Regardless of the input image dimensions, the compression branch receives compression information from original resolution images having a size $16 \times 128 \times 1$. Compression features extracted from original resolution images are standardized to have a mean value of 0 and a standard deviation value of 1 (zero-centered input) to aid in faster model convergence. The input size of $16 \times 128 \times 1$ helps keep the proposed framework 'lightweight,' i.e., the architecture processing input of this size need not be massive.

Two different architectures are employed as the compression branch in several variants of the proposed framework (Table 2). Besides trying a simple CNN, an involution neural network (INN) containing the novel involution kernel [24] (introduced in CVPR 2021) is also used as the compression branch. The novel involution-based neural network can achieve competent classification results at a lesser computational cost than CNN compression branch models. Fig. 3 demonstrates the feature extraction methodology of the two kernels.

Both the CNN and INN variants contain four convolution/involution layers. Each layer is accompanied by batch normalization. Max Pooling is used to reduce the dimensionality of feature maps. ReLU activation function is used. INN models specialize in reducing the number of trainable parameters while achieving competent results. Despite the fewer parameters in the INN compression branch, it can achieve competent classification results. The compression branch is also designed to produce the final features of size 16.

**3.4 Final Model**

The final model fuses the 16 features produced by each spatial and compression branch. It utilizes two fully connected layers to scale down the fused features to the final two features representing binary classification scores. ReLU activation and batch normalization are applied to the first fully connected layer output.

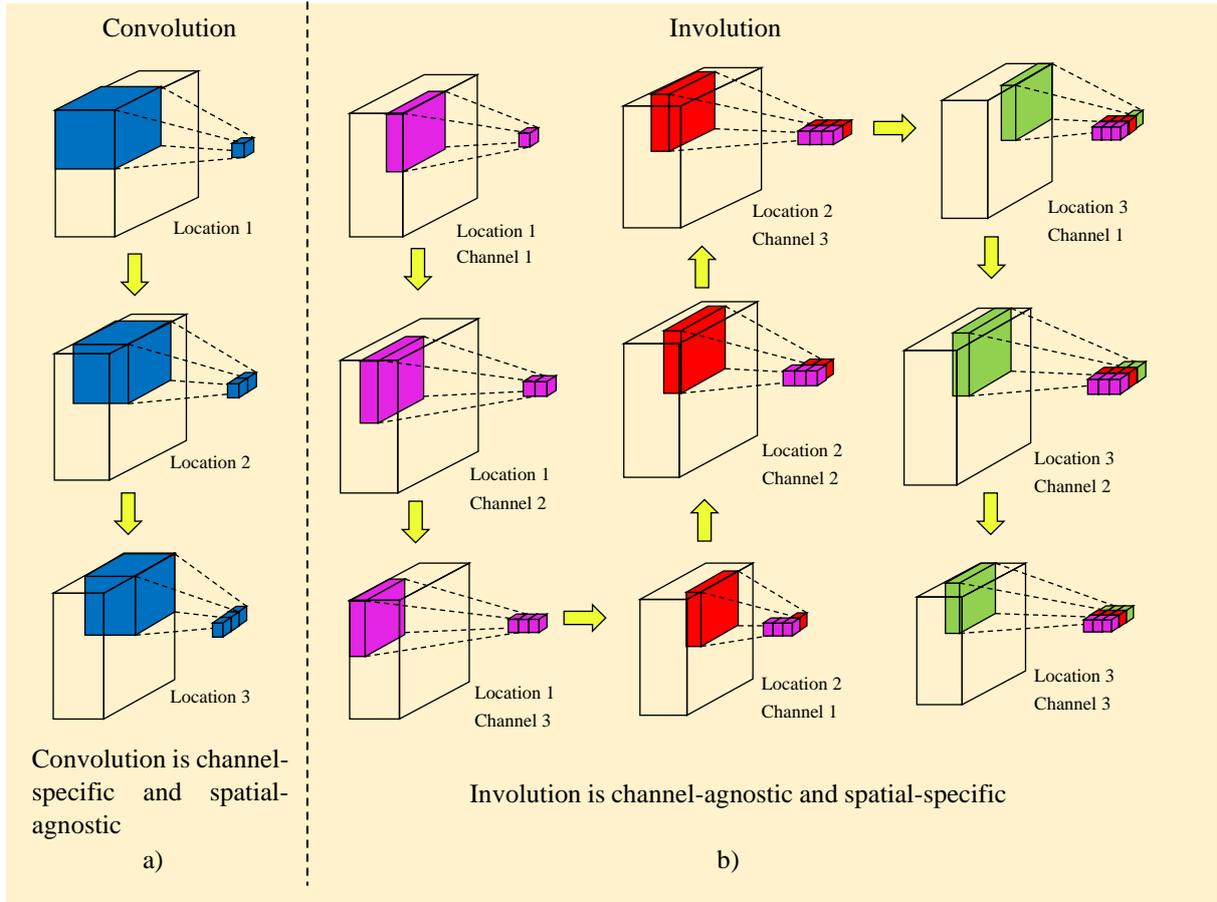

Fig. 3 Comparison of Convolution and Involution kernels. a) The number of channels in a convolution kernel is always equal to that of input. The convolution kernel (shown is blue) perfectly overlaps with input channel-wise and its kernel coefficients remain same for all strided locations (same blue color for all locations). b) The involution operator is a single channel kernel. Its coefficients remain same for all channels of input at a fixed spatial location. Unlike convolution, involution coefficients change with each strided location. Colors magenta, red and green represent involution operation at different spatial locations.

## 4 Experimental Setup

This section describes the experimentation conducted on the proposed splice detection framework. It describes the datasets, hardware, and performance metrics, training parameters for the proposed framework, results obtained, and a comparison of the proposed framework against existing state-of-the-art methods.

### 4.1 Datasets

One of the critical challenges in splice detection is the lack of large-scale splice datasets. Table 4 provides a list of existing splice detection datasets. Most of the existing splice datasets are limited in terms of the number of samples, and not all include binary masks for localization implementations. Training deep models on small datasets inevitably presents the problem of overfitting.

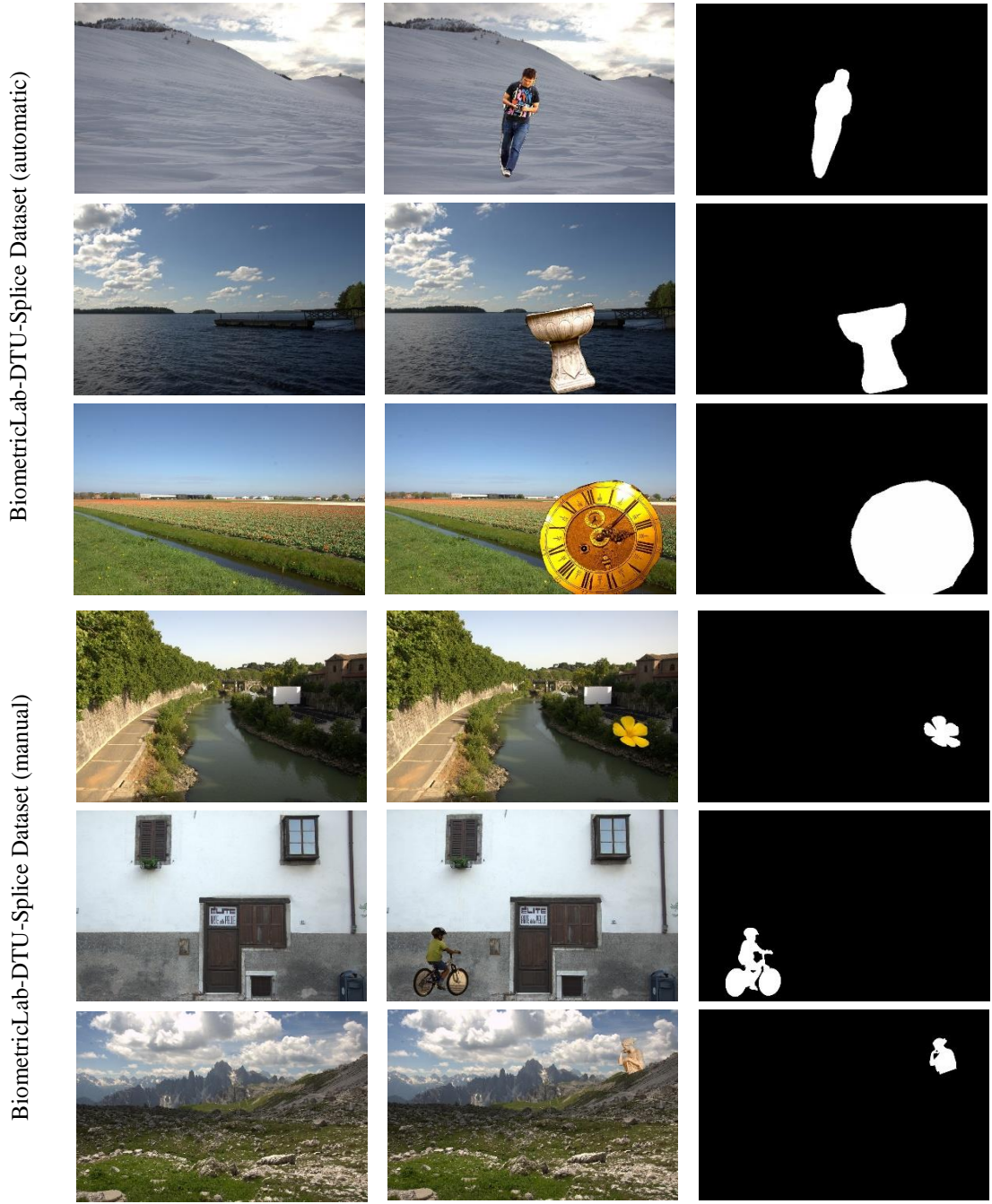

Fig. 4 Samples from the proposed BiometricLab-DTU Splice dataset

To this end, a novel splice detection dataset having two variants is proposed – BiometricLab-DTU-Splice Dataset (Table 3). The first variant (automatic) is autogenerated from the code. The second variant contains spliced images prepared in Adobe Photoshop Software.

Table 3 demonstrates the number of samples in each dataset used in this experiment. Fig. 4 demonstrates the proposed dataset's original, spliced, and binary mask images.

Table 3 Details of Proposed Datasets

| Dataset | Type | Original Samples | Spliced Samples | Total Samples | Splice Mask |
|---|---|---|---|---|---|
| BiometricLab-DTU-Splice dataset (automatic) | Proposed | 8156 | 8156 | 16312 | Yes |
| BiometricLab-DTU-Splice dataset (manual) | Proposed | 3106 | 3106 | 6212 | Yes |

| Dataset | Type | Original Samples | Spliced Samples | Total Samples | Splice Mask |
|---|---|---|---|---|---|
| CASIA v2.0 (modified) | Publicly Available | 5123 | 5123 | 10246 | - |

Table 4 Comparison of proposed splice detection dataset with existing splice datasets. The proposed dataset is mentioned in bold

| Ref. | Year | Dataset | Tampering Type | Original Samples | Spliced Samples | Resolution | Format | Splice Masks |
|---|---|---|---|---|---|---|---|---|
| **[27]** | 2004 | Columbia Gray | Splicing | 933 | 912 | 128 x 128 | BMP | |
| **[28]** | 2006 | Columbia Color | Splicing | 183 | 180 | 757 x 568, 1152 x 768 | TIFF | Yes |
| **[29]** | 2009 | CASIA v1.0 | Splicing | 800 | 921 | 384 x 256 | JPG | No |
| **[29]** | 2009 | CASIA v2.0 | Splicing | 7491 | 5123 | 240 x 160, 900 x 600 | TIFF, BMP, JPG | No |
| **[30]** | 2013 | DSO-I | Splicing | 100 | 100 | 2048 x 1536 | PNG | - |
| **[30]** | 2013 | DSI-I | Splicing | 25 | 25 | Variable | - | - |
| **[31]** | 2014 | Image Forensic Dataset Challenge | Splicing | 144 | 144 | 2018 x 1536 | PNG | - |
| **[32]** | 2015 | SYSU-OBJFORG dataset | Copy Move, Splicing | 100 | 100 | 1280 x 720 @ 25fps | H.264 / AVC | - |
| **[33]** | 2015 | GRIP | Splicing | 80 | 80 | 1024*768 | JPG | Yes |
| - | 2022 | **Biometric-Lab-DTU Splice Dataset (automatic)** | Splicing | 8156 | 8156 | 3008 x 2000, 4288 x 2848, 4928 x 3264 | JPG | Yes |
| - | 2022 | **Biometric-Lab-DTU Splice Dataset (manual)** | Splicing | 3106 | 3106 | 3008 x 2000, 4288 x 2848, 4928 x 3264 | JPG | Yes |

*4.1.1 BiometricLab-DTU-Splice Dataset (Automatic)*

The 'automatic' variant of the BiometricLab-DTU Splice dataset is autogenerated through Python code. 8156 high-resolution uncompressed images from the RAISE dataset [34] are used as source images. Several existing splice detection approaches have prepared datasets by compressing images with quality factors of step size 5 [26], [14]. The jpeg images of the proposed dataset are produced with a random integer quality factor which ensures a richer distribution of compression information compared to the above-described case.

*Dataset Generation of Automatic Spliced Images:* A pre-trained RCNN model trained on the MS-COCO dataset extracts objects from the original images of the RAISE dataset. This is achieved from the instance_segmentation() function of the PixelLib python library. This function extracts different objects detected as the result of segmentation. Next, the binary masks are generated for each object by converting the object images to grayscale and thresholding into binary images. The extracted objects are tampered with several random manipulations, including rotation, scaling, flipping, contrast changes, brightness variations and sharpness modifications before pasting onto another original image. The degree of scaling, rotation and other manipulations is random as shown in Fig. 5. Finally, the extracted objects are pasted onto original images using the paste() function of the Python's PILLOW library. Lastly, the spliced images are saved as jpeg images with random compression quality factor as shown in the "second compression" column of Fig. 5. The image reading and color-scale modifications are done using the open-cv library, while the rotation, scaling, flipping, contrast and sharpness enhancements are achieved with the PILLOW library. The binary masks produced from code can aid in future splice localization methods.

| name | first compression | second compression | object name | scale | rotation | flip | contrast | brightness | sharpness |
|---|---|---|---|---|---|---|---|---|---|
| r00816405t_spliced.jpg | 94 | 73 | object16_r0fb0a69( | 1 | 28 | 0 | 1.73 | 1.29 | 1.81 |
| r00869422t_spliced.jpg | 53 | 61 | object2_rccf44639t | 0.87 | 132 | 1 | 1.81 | 1.16 | 1.81 |
| r00879054t_spliced.jpg | 87 | 75 | object2_rb6c0981e' | 0.94 | 109 | 1 | 1.82 | 1.29 | 1.55 |
| r01058910t_spliced.jpg | 63 | 53 | object11_r854fa219 | 0.94 | 110 | 0 | 1.81 | 1.32 | 1.75 |
| r01170470t_spliced.jpg | 35 | 53 | object3_r39a8b2f2t | 1 | 69 | 1 | 1.74 | 1.35 | 1.69 |
| r01474187t_spliced.jpg | 51 | 34 | object3_r61184440 | 1 | 176 | 1 | 1.58 | 1.16 | 1.95 |
| r0150031ft_spliced.jpg | 74 | 63 | object5_r319e1687 | 1 | 149 | 1 | 1.65 | 1.27 | 1.86 |

Fig. 5 Parameters used during creation of spliced samples. Each pasted object undergoes five modifications namely scaling, rotation, flip, contrast and sharpness specified by randomized values. Each spliced image is compressed twice with random quality factors as shown in the 'first compression' and 'second compression' columns.

8156 high-resolution uncompressed images from the RAISE dataset were used to produce 8156 original and 8156 spliced images, which form the BiometricLab-DTU Splice dataset (automatic). Fig. 4 shows some samples from the proposed automatic dataset variant.

*4.1.2 BiometricLab-DTU-Splice Dataset (Manual)*

The second variant of the proposed dataset has been prepared manually. A total of 3106 images were spliced manually along with the same number of the original counterparts to formulate a balanced dataset of 6212 images. Spliced samples from the manual variant of the proposed dataset are more realistic visually than the automatic version due to the random manipulations on pasted objects of computer-generated spliced images in the automatic variant dataset. Table 4 compares the proposed BiometricLab-DTU Splice dataset against existing splice datasets. Most existing datasets fall short in terms of the total number of samples available, have smaller resolution images, and usually don't have splice masks to aid splice localization approaches when compared against the proposed dataset variants.

*Dataset Generation of Manually Spliced Images:* The source and target images are opened in the Adobe Photoshop software. An object is selected from the source image using the 'Quick Selection Tool'. The selected object is pasted onto the target image as a new layer. Resizing, rotation, flipping, contrast and brightness modifications are made randomly to the object layer. The pasted object is selected again and the 'Layer Mask Tool' in Photoshop generates the binary mask layer. Next, a 'Photoshop Action' is created that automates the process of saving the spliced image and its corresponding binary mask into separate folders. Specifically, the spliced image is saved as a jpeg image, while the binary mask is generated from the mask layer formed by the layer mask tool. For each image, opening the source and target images, selecting the object to be pasted and pasting operations are done manually. Then, the Photoshop action helps to save the spliced images and corresponding binary masks automatically.

*4.1.3 A modified CASIA v2.0 dataset*

The CASIA v2.0 is a challenging image tampering dataset containing 7491 original and 5123 tampered images [29]. However, over half of the tampered images are uncompressed (TIFF or BMP format). Since the proposed splice detection framework utilizes compression artifacts of image samples, the uncompressed format images were compressed with a random quality factor. This modified variant of the CASIA v2.0 dataset is used for experimentation.

Since all three datasets are novel (two proposed and one modified publicly available dataset), a comparison of the proposed framework with existing splice detection approaches is conducted by training all architectures (proposed and existing models) on these new datasets instead of merely comparing with metric figures mentioned in published works. 5123 original images are chosen from the original CASIA v2.0 dataset for experimentation to maintain class balance.

### 4.2 Hardware Resources and Evaluation Metrics

All experiments have been implemented using the Pytorch Library. All experiments are run on a system with 128 GB RAM and a 24 GB NVIDIA TITAN RTX graphic card. The metrics used to evaluate the proposed splice detection framework are *Accuracy* (ACC), *Precision* (P), *Recall* (R), *F1-score* (F1), *Area Under Curve* (AUC), and *Mathews Correlation Coefficient* (MCC).

Correctly identified spliced samples are counted as true-positive ($\mathbb{TP}$), original images correctly classified as original are counted as true-negative ($\mathbb{TN}$), original images misclassified as spliced are counted as false-positive ($\mathbb{FP}$) and spliced images misclassified as original are counted as false-negative ($\mathbb{FN}$). The various metrics are defined using Eq. (1)-(5).

$$Accuracy = \frac{\mathbb{TP}+\mathbb{TN}}{\mathbb{TP}+\mathbb{TN}+\mathbb{FP}+\mathbb{FN}} \tag{1}$$

$$Precision = \frac{\mathbb{TP}}{\mathbb{TP}+\mathbb{FP}} \tag{2}$$

$$Recall = \frac{\mathbb{TP}}{\mathbb{TP}+\mathbb{FN}} \tag{3}$$

$$\mathbb{F}1\ Score = 2 * \frac{Precision * Recall}{Precision + Recall} \tag{4}$$

$$MCC = \frac{\mathbb{TP}*\mathbb{TN} - \mathbb{FP}*\mathbb{FN}}{\sqrt{(\mathbb{TP}+\mathbb{FP})(\mathbb{TP}+\mathbb{FN})(\mathbb{TN}+\mathbb{FP})(\mathbb{TN}+\mathbb{FN})}} \tag{5}$$

### 4.3 Experimental Setup

This section describes the experimental setup used to evaluate the proposed splice detection framework.

*4.3.1 Dataset Variants, Pre-processing, and Data Augmentation*

The datasets mentioned in the previous section are combined, as shown in Table 5, to produce a larger pool of diverse input data. The AMC variant combines the proposed and modified CASIA datasets to produce one large dataset of 32770 images. The AM variant combines both variants of the proposed dataset to make a more extensive set of 22524 images. All variants of the proposed splice detection framework and existing splice detection architectures are trained and evaluated on these variants to ensure comprehensive training and evaluation in this experiment.

Table 5 Dataset Variants that are used for the experimentation.

| Dataset Variant | Code | Total Samples | Train & Validation Samples | Test Samples |
|---|---|---|---|---|
| BiometricLab-DTU-Splice dataset (automatic + manual) | AM | 22524 | 90% | 10% |
| All three datasets combined | AMC | 32770 | | |

Images are resized to $256 \times 256$ for all spatial models except Vision Transformer, which expects an input size of $384 \times 384$. Images are standardized to have a mean value of 0 and a standard deviation of 1 (channel-wise) and hence are 'zero-centered.' Three data augmentations are used for experimentation: random horizontal flip, random vertical flip, and random rotation from 0 to 180 degrees.

*4.3.2 Model Variants*

The models specified in Table 2 are combined and evaluated to verify the validity of the proposed splice detection framework. The combined architectures are named after their constituent branch models.

The ResNet-CNN variant uses a pre-trained resnet18 in the spatial branch and a simple CNN for the frequency branch. The GoogleNet-CNN variant has a GoogleNet architecture as its spatial branch and CNN as its frequency branch. Similarly, VGG-CNN, DenseNet-CNN, and ViT-CNN include VGG, DenseNet161, and Vision Transformer models in their spatial and CNN-based frequency branches.

The involution operator is also used in the frequency branch with fewer trainable parameters than the simple CNN-based frequency branch. Hence each of the above-mentioned spatial branch models is combined with an involution-based frequency branch (INN) to produce five more variants denoted by ResNet-INN, GoogleNet-INN, VGG-INN, DenseNet-INN, and ViT-INN.

Hence, ten model variants of the proposed splice detection framework are trained and evaluated in this experiment. The spatial branch of each variant is loaded with ImageNet weights and frozen except for the last layer.

*4.3.3 Hyperparameter Settings*

All experiments are run for 30 epochs. The batch size is 256 for all models except Vit-CNN and the train-test splits are set to 90% and 10%, respectively. Five-fold cross-validation is implemented to ensure comprehensive training. Several optimizers are tried, and the Adam optimizer consistently produces the best results. The learning rate (LR) is decayed linearly. Table 6 shows the hyperparameter settings for each model variant to obtain the best results.

Table 6 Hyperparameter setting for various variants of the proposed Splice Detection Framework

| Models | Epochs | Batch Size | Initial LR | LR Decay Factor | Step Size of LR Decay (epochs) | Optimizer |
|---|---|---|---|---|---|---|
| VGG-CNN | 30 | 256 | 0.001 | 0.5 | 1 | Adam |
| VGG-INN | 30 | 256 | 0.01 | 0.9 | 1 | Adam |
| ResNet-CNN | 30 | 256 | 0.001 | 0.5 | 2 | Adam |
| ResNet-INN | 30 | 256 | 0.01 | 0.9 | 1 | Adam |
| GoogleNet-CNN | 30 | 256 | 0.001 | 0.5 | 2 | Adam |
| GoogleNet-INN | 30 | 256 | 0.01 | 0.9 | 1 | Adam |
| DenseNet-CNN | 30 | 256 | 0.001 | 0.5 | 2 | Adam |
| DenseNet-INN | 30 | 256 | 0.01 | 0.5 | 2 | Adam |
| ViT-CNN | 30 | 576 | 0.001 | 0.5 | 1 | Adam |
| ViT-INN | 30 | 256 | 0.01 | 0.9 | 1 | Adam |

The CNN frequency branch-based architectures achieved their best results with an initial learning rate of 0.001 and decayed by 50% after every one or two epochs. The INN frequency branch-based architectures showed their best results when the learning rate was initialized to a higher value and decayed by a smaller factor. Specifically, the initial learning rate for INN-based models was 0.01 and decayed by 10% only (except for DenseNet-INN) after every one or two epochs.

**4.4 Performance of Proposed Splice Detection Framework**

This section presents the results of all model variants of the proposed splice detection framework.

*4.4.1 Results Obtained*

ACC, P, R, F1, AUC, and MCC scores obtained by model variants of the proposed framework are mentioned in Table 7.

Table 7 Performance of the proposed Splice Detection Framework in terms of Accuracy (ACC), Precision (P), Recall (R) and F1-score (F1), Area Under Curve (AUC), and Mathews Correlation Coefficient (MCC). Best scores are highlighted in bold

| Proposed Model Variants | Dataset | Input Size (Spatial Branch) | Trainable Parameter Count | ACC | P | R | F1 | AUC | MCC |
|---|---|---|---|---|---|---|---|---|---|
| VGG-CNN | AMC | 256 x 256 x 3 | 99,538 | 0.9305 | 0.9471 | 0.9116 | 0.9290 | 0.9749 | 0.8617 |
| VGG-INN | AMC | 256 x 256 x 3 | 76,854 | 0.9018 | 0.9125 | 0.8847 | 0.8984 | 0.9640 | 0.8038 |
| ResNet-CNN | AMC | 256 x 256 x 3 | 42,194 | 0.9379 | 0.9510 | **0.9208** | 0.9357 | 0.9749 | 0.8761 |
| ResNet-INN | AMC | 256 x 256 x 3 | 19,510 | 0.8571 | **0.9814** | 0.7328 | 0.8391 | 0.9472 | 0.7402 |
| GoogleNet-CNN | AMC | 256 x 256 x 3 | 50,386 | 0.9342 | 0.9458 | **0.9208** | 0.9331 | 0.9781 | 0.8687 |
| GoogleNet-INN | AMC | 256 x 256 x 3 | 27,702 | 0.9006 | 0.9384 | 0.8574 | 0.8961 | 0.9647 | 0.8042 |
| DenseNet-CNN | AMC | 256 x 256 x 3 | 69,330 | **0.9382** | 0.9578 | 0.9185 | **0.9378** | **0.9802** | **0.8772** |
| DenseNet-INN | AMC | 256 x 256 x 3 | 46,646 | 0.8779 | 0.9105 | 0.8278 | 0.8672 | 0.9431 | 0.7577 |
| ViT-CNN | AMC | 384 x 384 x 3 | 46,290 | 0.9376 | 0.9571 | 0.9144 | 0.9353 | 0.9790 | 0.8759 |
| ViT-INN | AMC | 384 x 384 x 3 | 23,606 | 0.8733 | 0.9209 | 0.8174 | 0.8661 | 0.9463 | 0.7516 |

The training loss and validation loss plots converge smoothly towards 0 and validation accuracy towards increases close to 1. The confusion matrix obtained after training each model variant confirms the strong classification capabilities of the proposed framework model variants.

*4.4.2 Result Analysis*

This section discusses the results obtained by model variants in the previous section. Table 7 presents the number of trainable parameters in each model.

The largest model is VGG-CNN with 99538 trainable parameters, while the lightest model is the ResNet-INN variant with only 19510 trainable parameters. This proves the lightweight nature of the proposed framework since deep learning models of moderate size easily contain a few million trainable parameters. In contrast, all variants of the proposed splice detection framework are incredibly lightweight.

The CNN frequency branch-based models consistently score more than 0.93 ACC, more than 0.92 F1-score, more than 0.97 AUC, and more than 0.86 MCC scores. The INN frequency branch-based models score more than 0.87 ACC (except for ResNet-INN), more than 0.86 F1-score F1 (except for ResNet-INN), more than 0.94 AUC, and more than 0.74 MCC. It is clear from the above scores that the reduced number of trainable parameters in the INN models resulted in a slight performance drop.

The DenseNet-CNN model achieved the best scores with 0.9382 ACC, 0.9378 F1, 0.9802 AUC, and 0.8772 MCC. ResNet-INN scored the highest precision score of 0.9814. ResNet-CNN and GoogleNet-CNN share the highest recall score of 0.9208.

Fig. 6 shows the ROC curves for each model variant of the proposed splice detection framework.

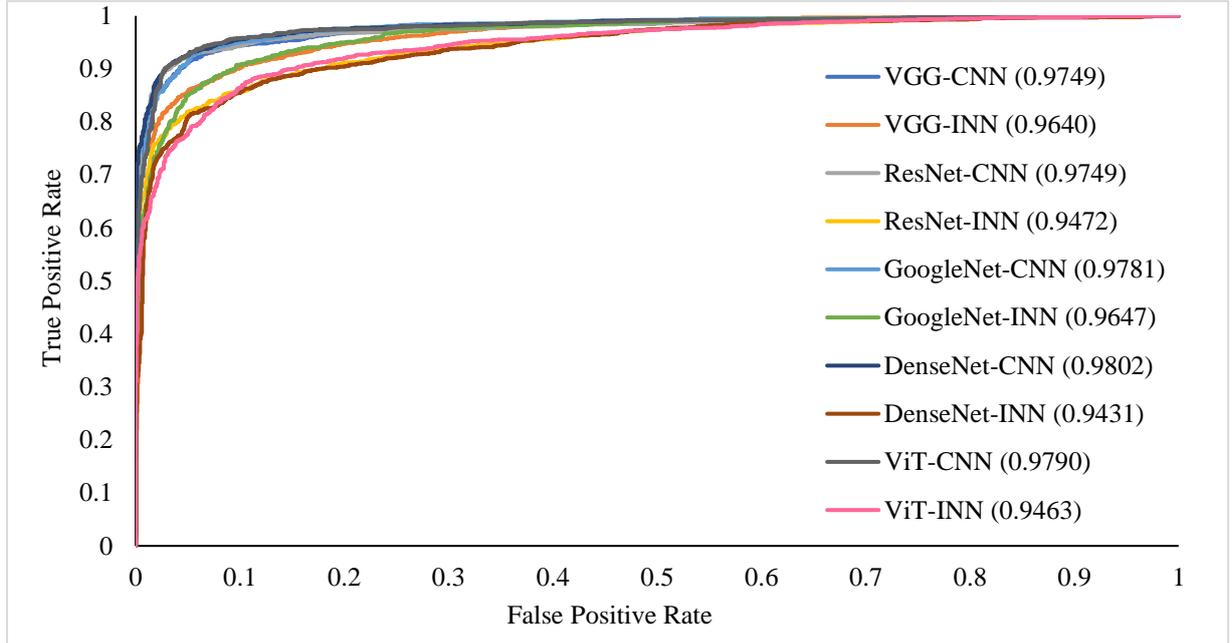

Fig. 6 ROC curves for different variants of the proposed framework. AUC scores are mentioned next to each variant name of the proposed Splice Detection Framework

*4.4.3 Ablation Study*

This section presents an ablation study of the proposed framework to confirm the validity of the contributions made by the spatial and compression branch. The spatial branch and frequency branch are trained and evaluated individually. Resnet18 is the spatial branch, loaded with ImageNet weights, and contains frozen layers except the last.

Table 8 Ablation study performance for the individual branches of the proposed Splice Detection Framework

| Models | ACC | P | R | F1 | AUC | MCC |
|---|---|---|---|---|---|---|
| Spatial Branch Only (ResNet) | 0.6957 | 0.7106 | 0.6495 | 0.6787 | 0.7602 | 0.3922 |
| Frequency Branch Only (CNN) | **0.7966** | **0.8376** | 0.7251 | **0.7773** | **0.8718** | **0.5971** |
| Frequency Branch Only (INN) | 0.6146 | 0.5786 | **0.8033** | 0.6727 | 0.7296 | 0.2523 |

Table 8 presents the results obtained from the ablation study, where each branch is trained on the same dataset as the proposed model. The best accuracy of 0.7966 is achieved by the CNN-based frequency branch, which is still very low compared to the performance of the proposed model variants. Similarly, the best F1, AUC, and MCC scores of 0.7773, 0.8718, and 0.5971, respectively, are very low. The scores in Table 8 indicate that the individual branches do not have high classification capability compared to the proposed model variants.

**4.5 Comparison with Existing Splice Detection Methods**

This section compares the proposed splice detection framework with existing state-of-the-art approaches. The comparison is based on classification metric scores and the size of models under comparison in terms of the number of trainable parameters.

*4.5.1 Comparison based on Classification Metrics*

This section presents the comparison of the proposed splice detection framework with the existing state-of-the-art methods for splice detection.

Table 9 presents a comparison of the proposed splice detection framework against existing state-of-the-art methods on the CASIA v2 dataset. The proposed framework outperforms all the mentioned approaches. The details of these comparison methods are given below.

- Zhang et al. [35] – Authors propose a multi-tasking dual-branch architecture for splice localization that learns from edge features of images and splicing masks. Both branches have encoder-decoder architecture for the splice localization purpose. Localization performance is improved by fusing features from shallow and deep layers of the model.

- Sun et al. [36] – Authors propose an edge-enhanced transformer model for splice localization. The transformer contains two branches to merge forgery clues with splicing edge clues. A novel feature enhancement module is used to highlight important features and supress noise.

- Yan et al. [37] – Authors combine the benefits of cross-attention and self-attention to propose a U-shaped transformer model that captures multi-scale spatial dependencies and finer contextual information. The transformer model integrates convolution, making it hybrid in nature.

- Bi et al. [13] – Author propose a novel 'Ringed Residual U-Net' model having residual feature propagation to solve the vanishing gradient problem. A residual feedback mechanism provides feature refinement charecteristics that widens the information from tampered and authentic regions within the image.

- Wu et al. [38] – Authors propose an end-to-end convolutional neural network based architecture for forgery detection and localization that can identify 385 image manipulations. Specifically, the proposed model contains a 'Image Manipulation Trace Feature Extractor' module to uncover forgery clues while another 'Local Anomaly Detection Network' module localizes the region of tampering.

- Chen et al. [39] – Authors design an end-to-end 'conditional random field' based convolutional network with region proposal network that focusses on small sized objects which are often missed by traditional neural networks. This helps in fine tuning the boundaries of spliced region during the localization of tampering.

- Liu et al. [40] – Authors utilize multiple convolutional networks to learn distinct image features at multiple scales. Then, conditional random fields are used to merge information from multiple scales to produce the final forgery localization results.

- Salloum et al. [9] – Authors proposed a dual-branch multi-tasking convolutional network that learns the surface label from one branch and identifies the edge of manipulated region from the second branch to localize tampering region.

- Wu et al. [41] – Authors highlight attentional manipulation traces with adaptive multi-scale fusion by using a 'content remove convolutional layer' that supresses the semantic content of images and highlights forgery clues. Another sub-network based on encoder-decoder design localizes forgery region.

- Chen et al. [42] – Authors combine 'multi-view feature learning' with 'multi-scale supervision' to learn generalized image tampering clues while keeping the false alarms over authentic images to a minimum. Multi-view feature learning exploits boundary clues around tampered region and noise distribution. Multi-scale supervision allows to uncover authetic image features to prevent false alarms.

- Xu et al. [44] – Authors utilize spatial features along with edge clues to identify image tampering. A novel 'Edge Gate' module helps to supress the semantic information of images and highlight tampering clues.

- Chen et al. [45] – Authors repurpose the tampering feature extration methodology. Instead of extracting clues from a two-stream Faster R-CNN model, authors design a novel two-layered convolutional structure to highlight forgery clues and hide image semantic details. The proposed module extracts weak forgery features crucial for tampering identification.

- Zhou et al. [46] – Authors propose a dual-stream Faster R-CNN based end-to-end model for forgery detection. The first stream extracts RGB clues for tampering such as contrast inconsistencies and uneven boundaries. The second stream extracts inconsistencies computed from steganalysis rich model filter layers.

- Wei et al. [47] – Authors improve the forgery detection scores by integrating an edge detection network with a Faster R-CNN model. The combined dual-modality information provides discriminative feature learning capability that excels in identifying image tampering.

Table 9 Comparison of Existing Splice Detection Approaches against the proposed splice detection framework on the CASIA dataset. The best scores are highlighted in bold. '-' represents scores that are not mentioned in the research papers.

| Model | ACC | P | R | F1 | AUC | MCC |
| --- | --- | --- | --- | --- | --- | --- |
| Zhang et al. [35] | - | - | - | 0.7653 | - | - |
| Sun et al. [36] | - | - | - | 0.6805 | - | - |
| Yan et al. [37] | - | 0.8090 | 0.7460 | 0.7350 | - | - |
| Bi et al. [13] | - | 0.6780 | 0.5860 | 0.5860 | - | - |
| Wu et al. [38] | - | 0.6310 | 0.6730 | 0.6510 | - | - |
| Chen et al. [39] | - | - | - | 0.7388 | - | - |
| Liu et al. [40] | - | - | - | 0.5232 | - | - |
| Salloum et al. [9] | - | - | - | 0.6675 | - | - |
| Wu et al. [41] | - | - | - | 0.5770 | - | 0.5590 |
| Chen et al. [42] | - | - | - | 0.6097 | - | - |
| Zhang et al. [43] | - | - | - | 0.6286 | - | - |
| Xu et al. [44] | - | - | - | 0.4601 | 0.8191 | - |
| Chen et al. [45] | - | 0.6616 | 0.7548 | 0.7051 | - | - |
| Zhou et al. [46] | - | 0.5044 | 0.6575 | 0.5709 | - | - |
| Wei et al. [47] | - | 0.5202 | 0.6642 | 0.5834 | - | - |
| **(Proposed) DenseNet-CNN** | 0.8125 | 0.7823 | 0.7646 | 0.7733 | 0.9126 | 0.6136 |
| **(Proposed) ResNet-CNN** | **0.8851** | **0.8123** | **0.9415** | **0.8664** | **0.9596** | **0.7744** |

Table 10 presents a comparison of the proposed framework on the combined dataset. Here again, the proposed framework performs better than other similar approaches. Four existing splice detection approaches have been implemented and evaluated in this section to demonstrate a fair comparison and illustrate the superiority of the proposed splice detection framework. The architectures implemented and the training process followed are the same as the research manuscripts mentioned.

Table 10 Comparison of Existing Splice Detection Approaches against the ResNet-CNN variant of the proposed splice detection framework on the combined dataset.

| Ref. | Model | Trainable Parameters | Dataset | ACC | P | R | F1 |
|---|---|---|---|---|---|---|---|
| [48] | (Existing) Dense CNN | 48,818 | AMC | 0.5850 | 0.5923 | 0.5498 | 0.5703 |
| [49] | (Existing) DCT + Quantization Table | 11,104,706 | AMC | 0.9273 | 0.9919 | 0.8951 | 0.9409 |
| [26] | (Existing) Multi-Domain CNN | 8,693,322 | AMC | 0.4693 | 0.4742 | 0.6413 | 0.5452 |
| - | (Proposed) DenseNet-CNN | 69,330 | AMC | 0.9382 | 0.9578 | 0.9185 | 0.9378 |
| - | (Proposed) ResNet-CNN | 42,194 | AMC | 0.9379 | 0.9510 | 0.9208 | 0.9357 |
| [17] | (Existing) Weighted Feature Fusion | 4,111,490 | AM | 0.5682 | 0.5837 | 0.4758 | 0.5245 |
| - | (Proposed) ResNet-CNN | 42,194 | AM | 0.9197 | 0.9508 | 0.9259 | 0.9382 |

The architecture in [48] contains four dense blocks with four, two, and two dense layers, respectively. Transition layers include 1 × 1 convolutions and Max pooling. Input images are converted to grayscale and resized to 256 × 256. Normalization is also done to the range of [0,1]. The model is trained for 50 epochs with an initial learning rate of 0.001, which is decayed by 10% every 1/3$^{rd}$ of an epoch. The optimizer used is SGD, and the batch size is 32.

The best architecture of [49], combining a quantization table to the last pooling and two fully connected layers, is implemented for comparison. The histogram range of Y channel DCT coefficients is [-60,60]. Train and test images are split by 90% and 10%, respectively. The model is trained for 50 epochs with a learning rate is 0.001 and an Adam optimizer. All convolution operations are accompanied by batch normalization.

The multi-domain CNN proposed in [26] is implemented and repurposed towards binary classification for image splice detection. The histogram range for DCT coefficients is [-50,50]. Train, validation, and test sets have sizes of 90%, 5%, and 5%, respectively. Both spatial and frequency branch use dropout and their respective inputs are applied with normalization to the range [0,1]. The model is trained for 50 epochs with the AdaDelta optimizer, with an initial learning rate of 0.01 is reduced by 10% every epoch.

The weighted feature combination architecture in [17], having four weight combination modules to combine YCbCr, Edge, and PRNU features, is implemented. The weight parameters $\propto_a, \propto_b$ and $\propto_c$ for each of the four weight modules are added to the computational graph and list of trainable parameters for automatic tuning during backpropagation. Cross-validation training is implemented with SGD optimizer, and the initial learning rate is 0.001, which decays by 10% every 10 epochs. Initial values for $\propto_a, \propto_b$ and $\propto_c$ are set to 0.3, 0.3 and 0.4, respectively.

PRNU features can be calculated from flat-field images of the source camera, or they can be estimated from a large number of natural images captured by a given camera device [50]. It is unclear how the authors computed PRNU features for CASIA dataset images since the dataset paper does not include source device information [29]. Hence, to alleviate this problem, the implemented weighted feature combination architecture is evaluated on the AM dataset variant, which includes the proposed Biometric-DTU-Splice dataset (automatic + manual). All images of the proposed Biometric-DTU-Splice dataset are derived from the RAISE dataset's uncompressed images whose camera device information is available. 50 images from each camera model ($N = 50$) are used to compute the PRNU factor $\widehat{K}$ using Eq. (6) for each camera device, as done in [50].

$$\widehat{K} = \frac{\sum_{k=1}^{N}(\mathcal{W}_k \mathcal{I}_k)}{\sum_{k=1}^{N}(\mathcal{I}_k)^2} \tag{6}$$

Here $\mathcal{I}_k$ is one of the multiple images from the source camera and $\mathcal{W}_k = \mathcal{I}_k - \widehat{\mathcal{I}_k}$ represents image noise residual. The implemented architecture is compared against the ResNet-CNN variant of the proposed framework, which is trained for a second time on the AM dataset variant for a fair comparison (Table 10).

The results from Table 10 indicate the superiority of the proposed 'lightweight dual-branch information preserving spatio-compression modal splice detection framework.' Only [49] of the existing splice detection methods can match the proposed framework's accuracy, precision, recall, and f1-scores. But it is an extremely heavy architecture with more than 11 million trainable parameters.

*4.5.2 Comparison based on Size (Number of Trainable Parameters)*

One of the design principles of the proposed splice detection framework is to make it 'lightweight.' This idea is ideally suited when the usual splice detection datasets are small and deep architectures are prone to overfitting. The proposed model reduces the number of trainable parameters by using transfer learning in the spatial branch and processing DCT features of size $16 \times 128 \times 1$ in the compression branch through extremely lightweight neural networks. The number of trainable parameters in each variant of the proposed splice detection framework is presented in Table 7 and that of existing splice methods in Table 10. Fig. 7 presents a size comparison of all the proposed splice detection framework variants (pink) and some existing splice detection methods (blue). The size difference (no. of trainable parameters) between the proposed and existing methods is so significant that a 'log-scale' is used to plot the size difference. Fig. 7 indicates the 'lightweight' nature of the proposed splice detection framework variants.

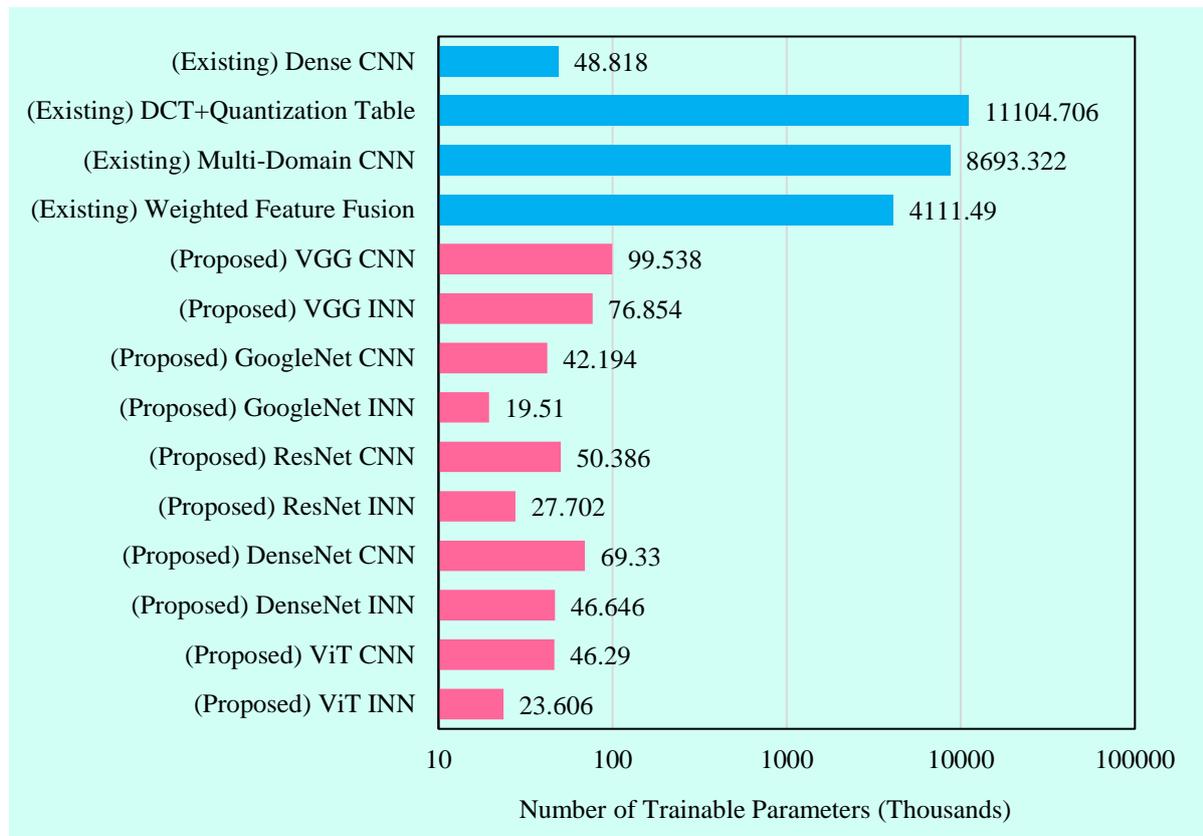

Fig. 7 Size comparison of the proposed Splice Detection Model Variants (pink) against Existing State-of-the-Arts (blue). Log scale is used to better demonstrate the massive differences in the number of trainable parameters.

## 5 Conclusion, Limitations & Future Work

### 5.1 Conclusion

This paper makes two-fold contributions towards splice detection in jpg images. Firstly, a novel splice detection dataset, the *'BiometricLab-DTU Splice dataset,'* is proposed. The proposed splice dataset has two variants: the first is autogenerated from code, and the second contains handmade spliced samples. The proposed splice detection dataset significantly adds to the existing small-scale splice datasets. Secondly, a novel 'Lightweight Dual-branch Information Preserving Spatio-Compression Modal Splice Detection Framework' is proposed that incorporates design principles consistent with today's splice detection research scenario (small-scale splice datasets). Several variants of the proposed framework are trained on images from the proposed splice dataset and a modified CASIA v2.0 dataset. Experimental results prove the superiority of the proposed splice detection framework compared to existing methods without requiring millions of trainable parameters in the neural network. A similar design principle can be used for future work to build a splice localization framework.

### 5.2 Limitations

Although the proposed framework is quite potent in identifying splice forgery in images, there are a few limitations.

*Firstly,* the proposed framework relies heavily on the jpeg compression artefacts as its compression branch inputs DCT compression information to make predictions. Hence, this framework works only for jpeg images. However, since data compression is widely used across all forms of multimedia data and jpeg compression is the most famous compression standard, the proposed framework is very relevant in present times.

*Secondly,* the proposed framework is designed specifically for image splice manipulations and is not trained for other manipulations such as copy-move. However, if the proposed framework is trained on multiple image manipulation types, it is likely to be highly effective due to its light-weight and information-preserving design.

## 5.3 Future Work

This section presents the possible future works based on the proposed splice detection framework and splice dataset created in this manuscript.

*Firstly,* the proposed splice dataset can be used to design and develop more robust splice detection models. Since, deep learning models require an abundance of labelled input data, the proposed splice dataset can help future models to achieve higher performance in both the detection and localization of splice manipulation in images.

*Secondly,* with the rise of transformer-based models in computer vision tasks, neural networks can now capture long-term visual dependencies often missed by convolution-based networks. These recent transformer models can be utilized to design newer, more capable models for image splice detection.

*Thirdly,* visual attention has become an integral part of vision models as they have consistently boosted model performances in various classification tasks. Visual attention can be integrated to boost model performance in identifying image splicing.

*Lastly,* generalized and unseen forgery detection models are still underexplored. There is a need to design models that can identify image tampering not present in the training images.